\documentclass[runningheads]{llncs}
\usepackage{multirow}
\usepackage{graphicx}
\begin{document}
% EchoTracker: Myocardial Tissue Tracking for Global Longitudinal Strain in Echocardiography
\title{EchoTracker: Advancing Myocardial Point Tracking in Echocardiography}
\titlerunning{EchoTracker: Point Tracking in Echocardiography}
% \author{Anonymous}
\author{Md Abulkalam Azad\inst{1}\orcidID{0009-0006-7177-4961} \and
Artem Chernyshov\inst{1} \and
John Nyberg\inst{1} \and
Ingrid Tveten\inst{1,3} \and
Lasse Lovstakken\inst{1} \and
Håvard Dalen\inst{1,2} \and
Bjørnar Grenne\inst{1,2} \and
Andreas {\O}stvik\inst{1,3}\orcidID{0000-0003-3895-2683}}
\authorrunning{M. A. Azad et al.}
%\authorrunning{Anonymous et al.}
% \institute{Anonymous Organization\\
% \email{**@******.**}
% }
\institute{Norwegian University of Science and Technology, Norway \and
Clinic of Cardiology, St. Olavs Hospital, Norway \and
SINTEF Digital, Norway\\
\email{\{md.a.azad,andreas.ostvik\}@ntnu.no}}
\maketitle              % typeset the header of the contribution
\begin{abstract}
Tissue tracking in echocardiography is challenging due to the complex cardiac motion and the inherent nature of ultrasound acquisitions. Although optical flow methods are considered state-of-the-art (SOTA), they struggle with long-range tracking, noise occlusions, and drift throughout the cardiac cycle. Recently, novel learning-based point tracking techniques have been introduced to tackle some of these issues. In this paper, we build upon these techniques and introduce EchoTracker, a two-fold coarse-to-fine model that facilitates the tracking of queried points on a tissue surface across ultrasound image sequences. The architecture contains a preliminary coarse initialization of the trajectories, followed by reinforcement iterations based on fine-grained appearance changes. It is efficient, light, and can run on mid-range GPUs. Experiments demonstrate that the model outperforms SOTA methods, with an average position accuracy of 67\% %with respect to reference%
and a median trajectory error of 2.86 pixels. Furthermore, we show a relative improvement of 25\% when using our model to calculate the global longitudinal strain (GLS) in a clinical test-retest dataset compared to other methods. This implies that learning-based point tracking can potentially improve performance and yield a higher diagnostic and prognostic value for clinical measurements %compared to%
than current techniques. Our source code is available at: https://github.com/riponazad/echotracker/.

\end{abstract}

\section{Introduction}
Myocardial imaging in echocardiography uses ultrasound (US) image analysis to assess and quantify the morphology and function of the cardiac muscle. These methods can be used to uncover reduced pumping efficiency, detect muscle irregularities, and diagnose various heart conditions, facilitating early identification of cardiac dysfunction. Myocardial strain, a measure of deformation, has shown high sensitivity with superior diagnostic and prognostic value compared to common anatomical measurements, such as ejection fraction~\cite{farsalinos2015head}. Motion estimation is vital for precise strain, but is hampered by variability in image acquisition, measurements, and inherent limitations of US. Currently, motion estimation and strain imaging rely on speckle tracking using block- and feature-matching approaches~\cite{voigt2015definitions}. %Advances in learning-based techniques, including optical flow-based architectures like FlowNet 2.0 \cite{ilg2017flownet} and PWC-Net \cite{sun2018pwc}, with modifications to adapt with US and strain calculations have enhanced these methods~\cite{evain2022motion,myhre2024external,ostvik2021myocardial}.%
Recent advances in learning-based techniques, such as optical flow-based architectures like FlowNet 2.0 \cite{ilg2017flownet} and PWC-Net \cite{sun2018pwc}, have inspired researchers to use and adapt them for US imaging and enhanced strain calculations~\cite{evain2022motion,myhre2024external,ostvik2021myocardial}.
However, optical flow estimates dense displacement fields between consecutive frames without considering long-range temporal context. This limitation makes tracking susceptible to noise, out-of-plane motion, and decorrelation of speckle patterns. Therefore, it hinders optimal tracking across multiple frames and causes drift within the cardiac cycle. 

In this study, we propose EchoTracker, a novel method for tissue tracking in echocardiography. It is designed based on the latest point tracking approaches for general applications \cite{doersch2022tap,doersch2023tapir,harley2022particle,karaev2023cotracker,zheng2023pointodyssey}. To the best of our knowledge, this is the first work that uses such an approach in the field of medical imaging. It addresses the limitations of dense optical flow by leveraging the temporal context of longer sequences. Herein, we build upon this and design our model architecture tailored for US data, enabling efficient learning with enhanced performance and a lightweight structure. We assess the tracking performance of our model compared to related approaches. Finally, we utilize our model for GLS calculations and compare the clinical performance with other SOTA solutions.

\section{Tracking Any Point (TAP)}
Tracking of arbitrary points in video sequences is a new research area in deep learning, evolving to mitigate the limitations of optical flow-based tracking. It also possesses the capability to tackle deformations when queried points are selected on the surface of non-rigid objects. Doersch et al. were the first to formalize the TAP problem, provided a benchmark comprising real and synthetic data, and proposed a simple baseline, TAP-Net, for evaluation \cite{doersch2022tap}. The TAP algorithm takes a video and query points from any frame $t$ as input and predicts locations $(x_t, y_t)$ and binary occlusions $(o_t)$ as output for each point in all other frames. Doersch et al. also mentioned that the output location $(x_t, y_t)$ is meaningless when occluded $(o_t=1)$. However, this poses a contradiction to our problem, as we still require the location to compute GLS even if the point is out of the plane. Our modifications to the formal definition of TAP align with Persistent Independent Particles (PIPs) \cite{harley2022particle}. It aimed to track particles as pixels across long-range video sequences, inspired by the classic Particle Video approach \cite{sand2008particle}.

However, PIPs has notable drawbacks, as it operates on long videos using temporal sliding, which can lead to drift when points are occluded for more than a single window. Expanding the temporal window makes the model slow and unsuitable for parallel computation. To this end, Doersch et al. later introduced TAPIR \cite{doersch2023tapir}, leveraging both TAP-Net and PIPs. Also in parallel, Zheng et al. proposed PIPs++ by modifying PIPs with expanded temporal receptive fields and a multi-template update mechanism \cite{zheng2023pointodyssey}.

Although these approaches differ, a common factor is that they all track points independently, not sharing information between trajectories. This limitation may hinder the tracking of deformable objects like myocardial tissue and lead to drift for points outside the US probe view. This issue has been addressed by CoTracker \cite{karaev2023cotracker} and OmniMotion \cite{wang2023omnimotion}. CoTracker iteratively refines trajectories using a transformer architecture in a sliding window manner after a naive initialization. Consequently, it shares the same disadvantage as PIPs for long-range tracking and also exponentially increases the time complexity in case of longer temporal windows. On the other hand, OmniMotion provides a test-time optimization approach based on the canonical 3D volume of the input video.

\section{Methods}
\begin{figure}
\centering
\includegraphics[width=0.95\textwidth]{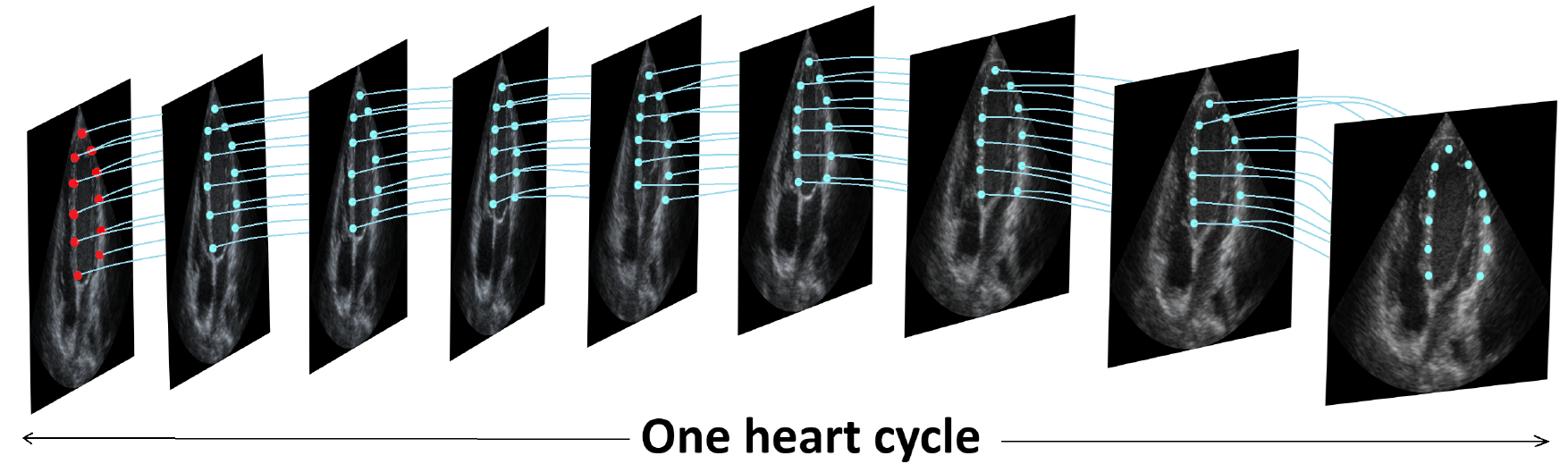}
\caption{An illustration of tracking queried points (highlighted in red) from the first frame throughout one heart cycle.} \label{ultra_tracker}
\end{figure}

The overall goal of our learning-based model is to track tissue points through the cardiac cycle while dealing with complex motion, deformation, and noise. As depicted in Fig.~\ref{ultra_tracker}, the input comprises a sequence of US images $U = \{u_s \in \bbbr^{H\times W}\}$ for $s=0, 1,..., S$, and a set of query points on the first frame $p_0=\{(x^n_0, y^n_0)\}$ for $n=0, 1,...,N$. Here, $H$ is the height of the image, $W$ is the width, while $x$ and $y$ are the horizontal and vertical pixel locations of a given point. The proposed solution outputs the locations of the queried points in all other frames, $P = \{p_s \in (x_s^n, y_s^n)\}$. Hence, the problem can be summarized as EchoTracker$(u_s, p_0) = p_s \in (x_s^n, y_s^n)$.

\subsection{EchoTracker}
\begin{figure}
\centering
\includegraphics[width=0.95\textwidth]{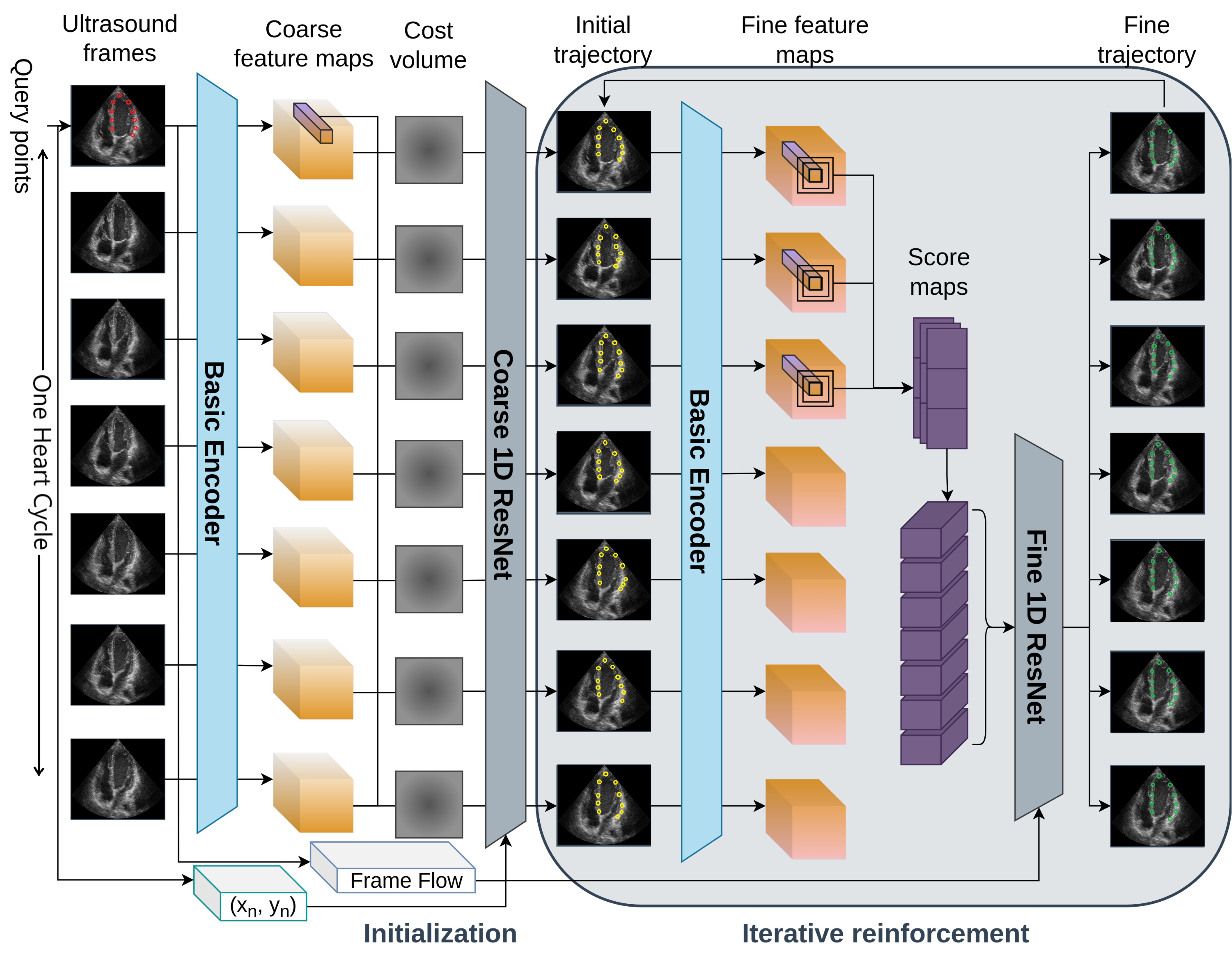}
\caption{EchoTracker is a two-fold coarse-to-fine model. Initially, it estimates coarse trajectories (yellow points) based on the cost volume for the given query points (red). It then imposes iterative reinforcement to obtain the fine trajectories (green points).} \label{echo_tracker}
\end{figure}

Our proposed model, named EchoTracker, includes two stages as shown in Fig.~\ref{echo_tracker}, \textit{initialization} and \textit{iterative reinforcement}. The approach follows a two-fold coarse-to-fine strategy inspired by TAPIR \cite{doersch2023tapir}. In the initial stage, trajectories are initialized based on the coarse resolution of the feature maps using a coarse network. Subsequently, in the second stage, the trajectories are iteratively refined using fine-grained feature maps by a fine network, thus constituting a two-fold coarse-to-fine approach. This technique not only speeds up computation but also prevents the loss of important information due to downsampling. Although the networks in both stages estimate trajectories independently, they exploit point locations from the first frame to maintain spatial correlation and estimate coherent trajectories. Additionally, frame flow (i.e. $u_s-u_{s-1}$), representing the difference between consecutive frames, is naively passed to the model to make it aware of global appearance changes. The model can run on US sequences of any length and with any number of query points, depending on available memory. 

\subsubsection{Initialization.} 
The input contains $S$ ultrasound images and $N$ number of query points, as highlighted by red dots in Fig.~\ref{echo_tracker}. We utilize a pruned 2D residual convolutional network (basic encoder)~\cite{he2016deep} to generate coarse feature maps $F_s \in \bbbr^{d\times \frac{H}{k}\times \frac{W}{k}}$ for each frame with $k=8$ and $d=64$. The pruning is motivated by a reduction in computational costs and the limited variability of data representation (e.g. grayscale, cyclic, velocity)  present in echocardiography. Given a query point in the first frame $p_0^n$, we extract a feature vector $f_{p_0^n} = \texttt{sample}(F_0, p_0^n)$, which captures the appearance of the point, through a bilinear sampling of the feature maps $F_0$ at that specific point location. Following that, we compute the cost volume $C_s^n = f_{p_0^n} \cdot \texttt{pyramid}(F_s)$ by correlating $f_{p_0^n}$ with features across the entire video, using multi-scale feature pyramids with $L=4$ levels and kernel size $r=3$. Finally, the cost volume is fed to a coarse 1D ResNet, similar to PIPs++~\cite{zheng2023pointodyssey}, which convolves across time to estimate the trajectory $P_s^n$ for $s=0,1,...,S$. This is highlighted in yellow in Fig.~\ref{echo_tracker}. We choose a 1D ConvNet here rather than a 2D to prioritize temporal information, assuming that bilinear sampling already perceives the required spatial information present in the US images.  

\subsubsection{Iterative Reinforcement.} 
Taking inspiration from PIPs~\cite{harley2022particle} and more recent point tracking methods, we also refine the initial coarse trajectories through an iterative reinforcement process illustrated in Fig.~\ref{echo_tracker}. We hypothesize that our strong initialization can substantially decrease the number of iterations necessary to converge to a refined trajectory, and thus keep it at $I=4$ for both training and evaluation. After initialization, we use the same basic encoder with a reduced downsampling factor $k=2$ to produce fine feature maps $F_s \in \bbbr^{d\times \frac{H}{2}\times \frac{W}{2}}$. Subsequently, unlike the initialization, we compute the cost volume $C_s^n = f_{p_s^n} \cdot \texttt{multicrop-pyramid}(F_s)$ for the query point by correlating the feature vector $f_{p_s^n}$ from the current frame $s$ with multi-scale crops of pyramid features on the fine feature maps surrounding the point location. Inspired by the feature extraction within a fixed temporal span of the current frame~\cite{azad2023multi} and multi-template update~\cite{zheng2023pointodyssey}, we track changes in the point appearance by acquiring additional cost volumes (i.e., $C_{s-2}^n, C_{s-4}^n$) at a fixed temporal span from the current frame and always for the first $C_0^n$. Afterwards, we concatenate these cost volumes and pass them through a linear layer to obtain the score maps for the current frame. Similarly, we obtain score maps for all frames and concatenate them before passing to the next layer to generate updates $\Delta p_s^n$ for the trajectory. This layer contains a fine 1D ResNet, which is a deeper and more weighted version of the coarse ResNet utilized in the initialization. Finally, updates are applied to the points to obtain the refined trajectory $p_{s,i}^n = p_{s,i-1}^n +\Delta p_{s,i-1}^n$, and the iterative reinforcement ensures the most fine-grained estimated trajectory. We supervise the model by considering all iterations in an end-to-end manner using the same loss function as in \cite{zheng2023pointodyssey}.

\section{Experiments and Results}
\subsection{Datasets}
Learning-based methods are usually trained using synthetic data and tested on real data annotated by humans~\cite{evain2022motion}. Unfortunately, the availability of high-quality synthetic point tracking data for echocardiography that can adequately replicate real-world data remains scarce. In this work, we chose to rely on human-annotated data, similar to real-world data annotations in TAP benchmarks~\cite{doersch2022tap}. Our trajectories are generated in a semi-supervised fashion using a traditional tracking algorithm and undergo tuning for optimal tracking by clinical experts. In addition,  experts perform quality assurance by rejecting points that do not follow the tissue properly, and we discard those from training and evaluation. We retrieve four splits, as illustrated in Table~\ref{tab_ds}, focusing on tracking the left ventricle myocardium from three acoustic views, namely the apical four-chamber, two-chamber and long-axis. All data were collected using GE Vivid E95 scanners, and received approval for research use from the regional ethics committee. DS-A is a test-retest dataset, meaning that the same patient is being scanned twice in immediate succession by two different operators. Therefore, we expect the patient to be in the same physical condition for both exams, serving as a reference for the reproducibility of the method. This dataset is independent and used exclusively for testing. On the other hand, DS-B, DS-C, and DS-D are subsets of the same dataset but analyzed by three different observers through random selection. We train the model using these datasets and evaluate its performance on the DS-A dataset.

% DS_A = HUNT4\_TEST\_RETEST
% DS_C = HUNT4\_NORMALS [JN]
% DS_D = HUNT4\_NORMALS [BG]
% DS_E = HUNT4\_NORMALS [HD]
\begin{table}
\centering
\caption{Ultrasound point tracking datasets and selected characteristics. The number of points and frames, as well as the height and width of the images, are given as average with range (min, max) in parenthesis.}\label{tab_ds}
\setlength{\tabcolsep}{3pt} % Adjust the space between columns
\begin{tabular}{lcccccccccc}
\hline
 {\textbf{Dataset}}& \textbf{Patients} & {\textbf{Videos}}& \textbf{Points} & \textbf{Frames} & \textbf{Height} & \textbf{Width}\\
\hline
 DS-A & 40 & 210 & 73 (57-91) & 93 (57-150) & 593 (583-640) & 571 (496-650) \\
 DS-B & 913 & 2731 & 81 (49-115) & 85 (44-151) & 592 (583-640) & 630 (453-844)\\
 DS-C & 615 & 1837 & 80 (49-111) & 84 (44-150) & 592 (583-640) & 631 (496-778)\\
 DS-D & 643 & 1922 & 66 (45-111) & 84 (46-151) & 592 (421-640) & 634 (392-856)\\
\hline
\\[-0.7cm]
\end{tabular}
\end{table}

\subsection{Implementation details}
Similar to Zheng et al.~\cite{zheng2023pointodyssey}, we use a learning rate of $5 \cdot 10^{-4}$, a one-cycle scheduler~\cite{smith2019super}, and the AdamW optimizer. Images are resized to $256 \times 256$ for both training and evaluation by default to limit the GPU memory footprint. Initially, we train our model on DS-B for 22 epochs using a sequence length of $S=36$. We fine-tune the model on DS-C for 50 epochs with $S=68$, and then on DS-D for a single epoch with $S=68$. Throughout the training process, we consistently use all available points with a batch size of 1. The training time on a single GPU, completing 50 epochs on the DS-B dataset typically takes over one week. We implemented our framework in PyTorch and used an NVIDIA GeForce RTX 3090 (24GB) GPU for both training and evaluation of models.
% After initially training on a shorter sequence (S=36), we fine-tune the model using the longest possible sequence, which is S=68. The number of training epochs usually ranges from 25 to 50 as even completing 50 epochs on the largest DS\_B dataset typically requires over a week.

\subsection{Evaluation}
\subsubsection{Evaluation Metrics.} Following SOTA point tracking literature, we report average position accuracy ($<\delta_{avg}^x$) as proposed in TAP-Vid~\cite{doersch2022tap} for the technical performance. Position accuracy ($<\delta^x$), measures the proportion (\%) of all the query points that fall within the threshold distance of pixels (e.g., $x=1, 2$) from their ground truth across the entire video and $<\delta_{avg}^x$ averages over five thresholds: 1, 2, 4, 8, and 16 pixels. Given the sensitivity of tracking in US, we also report $<\delta^x$ for individual threshold and median trajectory error (MTE)~\cite{zheng2023pointodyssey}, which calculates the distance in pixels between the estimated and ground truth trajectories. To show the efficiency of our model, we also include the average inference time (AIT) per video measured in seconds. Furthermore, to evaluate our model in a clinical setting, we calculate the peak GLS, which is defined as the relative change of the longitudinal ventricular length from end-diastole to minimum ventricular length.

% However, as $p_0$ is given as an input, it should be excluded from the output during the evaluation of the model. 

\subsubsection{Technical Performance.} 
For reference, we conduct an evaluation of current SOTA models directly on our DS-A dataset without fine-tuning. These results are summarized in Table~\ref{tab_tech_oob} which shows that PIPs++ and CoTracker stand out. Thus, we use these two models as baselines for our subsequent experiments.

\begin{table}
    \centering
    \caption{Performance comparison of the state-of-the-art methods on the DS-A dataset without fine-tuning.}
    \setlength{\tabcolsep}{7pt} % Adjust the space between columns
    \begin{tabular}{lccccccc}
    \hline
    \textbf{Method} & $<\delta^1\uparrow$ & $<\delta^2\uparrow$ & $<\delta^4\uparrow$ & $<\delta^8\uparrow$ & $<\delta^{16}\uparrow$ & $<\delta^x_{avg}\uparrow$ & MTE$\downarrow$\\
    \hline
    PIPs & 5 & 12 & 28 & 54 & 81 & 36 & 10.22\\
    TAP-Net & 5 & 17 & 41 & 70 & 89 & 44 & 6.50\\
    TAPIR & 7 & 21 & 47 & 78 & 95 & 50 & 5.63\\
    PIPs++ & 8 & 22 & 47 & 78 & \textbf{96} & 50 & 5.31\\
    CoTracker & \textbf{13} & \textbf{27} & \textbf{52} & \textbf{79} & 95 & \textbf{53} & \textbf{5.28}\\
    \hline
    \end{tabular}
    \label{tab_tech_oob}
    \\[-0.1cm]
    \raggedright
    {\scriptsize ~ $\delta$: Position accuracy (\%), MTE: Median trajectory error (pixel)}
    \\[-0.3cm]
\end{table}

\noindent The performance of EchoTracker compared with PIPs++~\cite{zheng2023pointodyssey} and CoTracker~\cite{karaev2023cotracker} fine-tuned with the same training datasets is displayed in Table~\ref{tab_tech_perf}. CoTracker shows a slight decline in performance after fine-tuning, likely attributed to suboptimal implementation details or data processing,  necessitating in-depth investigation in future studies. Our model demonstrates superior performance across all metrics, surpassing other methods by a significant margin. Moreover, it shows faster inference times compared to these alternatives. 

% Note that we include the models without fine-tuning solely to demonstrate the extent of improvement achieved from the initial stage, rather than for direct comparison purposes.

\begin{table}
    \centering
    \caption{Technical performance of EchoTracker on the DS-A test-retest dataset compared to state-of-the-art methods.}
    \setlength{\tabcolsep}{3pt} % Adjust the space between columns
    \begin{tabular}{lcccccccc}
    \hline
    \textbf{Method} & $<\delta^1\uparrow$ & $<\delta^2\uparrow$ & $<\delta^4\uparrow$ & $<\delta^8\uparrow$ & $<\delta^{16}\uparrow$ & $<\delta^x_{avg}\uparrow$ & MTE$\downarrow$ & AIT$\downarrow$\\
    \hline
    PIPs++ & 15 & 36 & 70 & 94 & \textbf{100} & 63 & 3.28 & 0.42\\
    CoTracker & 8 & 22 & 47 & 78 & 96 & 50 & 5.45 & 1.67\\
    EchoTracker (ours) & \textbf{19} & \textbf{43} & \textbf{76} & \textbf{96} & \textbf{100} & \textbf{67} & \textbf{2.86} & \textbf{0.24}\\
    \hline
    \end{tabular}
    \label{tab_tech_perf}
    \\[0.1cm]
    \raggedright
    {\scriptsize ~ $\delta$: Position accuracy (\%), MTE: Median trajectory error (pixel), AIT: Average inference time (s)}
    \\[-0.1cm]
\end{table}

\noindent To investigate if our two-stage architecture can reduce the number of iterations in the reinforcement stage, we solely train the initialization part on a limited part of the DS-B dataset. This initialization part yields $< \delta^x_{avg} = 48\%$, surpassing the performance of baseline PIPs and TAP-Net. Thus, it can be expected that the iterative reinforcement stage would require less effort to refine and smooth the initial trajectory. Our empirical investigations found that models trained on short sequences struggle when tracking longer videos. Therefore, a straightforward improvement was to train on longer sequences of one heart cycle. Furthermore, we experiment with replacing the fine ResNet with a transformer, following the approach in CoTracker \cite{karaev2023cotracker}. Surprisingly, this modification led to a drop in performance. The reason may be attributed to the fact that transformers typically require pretraining on large-scale datasets to outperform ConvNets~\cite{dosovitskiy2021an}. Exploring this aspect further could be an intriguing avenue for future research as we witness CoTracker outperform all the other methods without fine-tuning.

\subsubsection{Clinical Performance.}
In Table~\ref{tab_clinic_perf}, we present the GLS results compared to the reference measurements and in a test-retest situation. We also list results from solutions developed by others, albeit calculated on their private datasets. Our method, as well as fine-tuned PIPs++, performs favourably compared to other published work. However, the methods are tested on different private datasets, so a direct comparison was not possible.

\begin{table}[h]
    \centering
    \caption{Clinical results for GLS calculations compared to reference measurements and in a test-retest scenario.}
    \setlength{\tabcolsep}{7pt} % Adjust the space between columns
    \begin{tabular}{lcccccccc}
    \hline
    \multirow{2}{*}{\textbf{Method}} & \multicolumn{3}{c}{\textbf{Reference}} & & \multicolumn{3}{c}{\textbf{Test-retest}} \\
    \cline{2-8}
    & $\mu$ & $\sigma\downarrow$ & $MAD\downarrow$ & & $\mu$ & $\sigma\downarrow$ & $MAD\downarrow$\\
    % \textbf{Method} & $\mu\downarrow$ & $\sigma\downarrow$ & $MAD\downarrow$\\
    \hline
    c-PWC-Net-60A~\cite{evain2022motion} & 1.85 & 2.73 & N/A & & & &   \\
    us2ai~\cite{myhre2024external} & 0.68 & 2.52 & 2.0 & &  &  &  \\
    EchoPWCNet~\cite{salte2023testretest,salte2021aistrain} & -1.4 & 1.9 & 1.8 & & \textbf{0.0} & 1.9 & 1.6 \\[0.2cm]

    PIPs++ & -1.21 & 1.95 & 1.76 & & 0.11 & 1.62 & 1.28 \\
    CoTracker & -0.82 & 2.40 & 1.98 & & -0.11 & 2.47 & 1.96 \\
    EchoTracker (ours) & \textbf{-0.13} & \textbf{1.78} & \textbf{1.36} & & -0.13 & \textbf{1.55} & \textbf{1.21} \\
    \hline
    \end{tabular}
    \label{tab_clinic_perf}
    \\[0.1cm]
    \raggedright
    {\scriptsize ~ $\mu$: Bias (\%), $\sigma$: Standard deviation (\%), $MAD$: Mean absolute deviation (\%)}
    \\[-0.6cm]
\end{table}

\section{Conclusion}
We have introduced modern general-purpose point tracking approaches in echocardiography. Through a comprehensive analysis of several SOTA methods, we design a two-fold coarse-to-fine architecture and propose a novel model for tracking myocardial tissue. Our assessment demonstrates that this new approach not only outperforms SOTA solutions but also enhances the measurements of GLS in a test-retest scenario. Thus, learning-based point tracking holds the potential to elevate both the diagnostic and prognostic utility of myocardial function measurements, marking a notable step forward in the field of echocardiography.  

\bibliographystyle{splncs04}
\bibliography{ref}
% \bibliography{ref}




% LNCS Homepage, \url{http://www.springer.com/lncs}. Last accessed 4
% \end{thebibliography}
\end{document}